%% file: SIGGRAPH.tex
\documentclass[sigconf,nonacm]{acmart}

\usepackage{booktabs} 

\usepackage[ruled]{algorithm2e} 

\SetAlFnt{\small}
\SetAlCapFnt{\small}
\SetAlCapHSkip{0pt}

\begin{document}
\title{TryOnGAN: Body-Aware Try-On via Layered Interpolation}

\author{Kathleen M Lewis}
\affiliation{%
  \institution{Massachusetts Institute of Technology \& Google Research}
 }
\email{kmlewis@mit.edu}
\authornote{Work done while the first author was an intern at Google Research.}

\author{Srivatsan Varadharajan}
\affiliation{%
  \institution{Google Research}
}
\email{srivatsanv@google.com}
\author{Ira Kemelmacher-Shlizerman}
\affiliation{%
 \institution{Google Research \& University of Washington}
 }
\email{kemelmi@google.com}




\begin{abstract}
Given a pair of images—target person and garment on another person—we automatically generate the target person in the given garment. Previous methods mostly focused on texture transfer via paired data training, while overlooking body shape deformations, skin color, and seamless blending of garment with the person. This work focuses on those three components, while also not requiring paired data training. We designed a pose conditioned StyleGAN2 architecture with a clothing segmentation branch that is trained on images of people wearing garments. Once trained, we propose a new layered latent space interpolation method that allows us to preserve and synthesize skin color and target body shape while transferring the garment from a different person. We demonstrate results on high resolution $512\times 512$ images, and extensively compare to state of the art in try-on on both latent space generated and real images. 
\end{abstract}

%
%


%
%


\begin{teaserfigure}
   \centering
   \includegraphics[width=\textwidth]{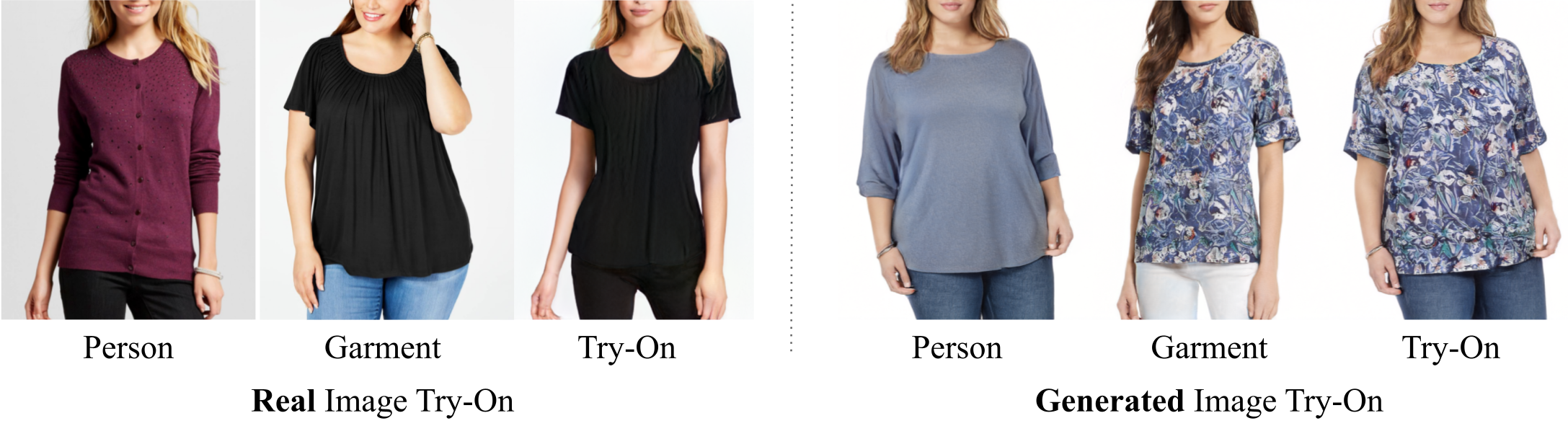}
   \caption{TryOnGAN is a StyleGAN-based interpolation optimization algorithm for photo-realistic try-on. Left: Shirt try-on for real images. Our method generates high quality synthesis of try-on images w.r.t. body shape, skin color, hair, and seamless blending and  warping the garment with the target person.  Right: Shirt try-on for latent space generated images. On generated images, our method is able to furthermore synthesize and correctly transfer high frequency details such as geometric patterns and complex textures. Zoom in to see the quality and details of the results.}
   \label{fig:teaser}
\end{teaserfigure}

\maketitle

\section{Introduction}

Virtual try-on—the ability to computationally visualize a garment of interest on a person of one’s choice—may become an essential part of an apparel shopping experience. A  useful  try-on, however, requires  high quality visualization, ideally indistinguishable from a photograph in a magazine with attention to body shape and type details. As a step towards this goal, we introduce a novel  image based try on algorithm, named \textit{TryOnGAN}, which seamlessly integrates person-specific components from one image with the garment shape and details from another image. Our experimental evaluation demonstrates state of the art photo-realistic results at the high resolution of $512\times512$ pixels.  
\begin{figure*}
    \centering \includegraphics[width=\textwidth]{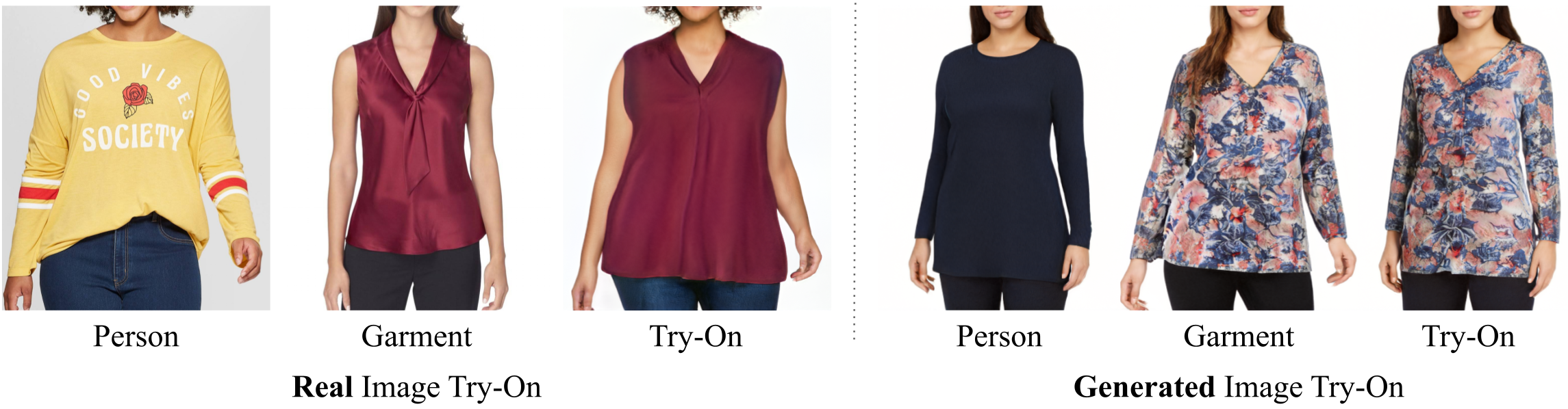}
    \caption{Results from our method on real and generated images. We run the same optimization method on both types of images, with real image try-on requiring an additional projection step to the latent space.  On real images, our method generates high resolution try-on results  that accurately transform for body shape and synthesize skin consistent with identity. However, some garment details are lost in projection. On generated images, our method can accurately transfer complex garment textures and patterns in addition to correctly synthesizing identity details, e.g., hair, body shape, skin color. While texture details projection for real images is left for future work, our algorithm already as is outperforming state of the art.}
    \label{fig:real-vs-generated}
\end{figure*}

We are motivated by the photo-realism and high resolution results of StyleGAN \cite{StyleGAN,StyleGAN2} for faces, and use it as our starting point for fashion try-on. We design a modified StyleGAN2 network, conditioned on 2D human body pose with a clothing segmentation branch. This architecture is used to train a model on 100K \textbf{unpaired} fashion images. During test time, given a pair of images—a person image and a garment image— we propose a method that automatically searches for  optimal interpolation coefficients \textit{per layer}, such that, when applied on the two images result in try-on. Interpolation coefficients are applied on the latent representations of the two input images, and are used to generate a single output image where the person from the first input image is wearing the garment from the second image. Figures \ref{fig:teaser} and \ref{fig:real-vs-generated} demonstrate results on both real and generated images. Real image try-on requires an additional projection step ahead of the core algorithm. 

Per layer optimization  enables semantically meaningful and high quality results. Unlike previous general GAN editing methods \cite{Edo, NoiseInjection}, which require \textbf{manual} choice of noise injection structure or manual identification of clusters and fixed parameters for all layers, our method \textbf{automatically} computes the best interpolation coefficients per layer by optimizing a loss function designed to preserve the identity and body of the person while warping  only the garment. Results section shows comparison to \citet{Edo} and the importance of per layer optimization.

Our method outperforms the state of the art \cite{ADGAN,CPVITON, ACGPN} with respect to three components: body shape, photorealism, and skin color preservation on real images, as well as generalizing to other datasets, while only using unpaired data.  While our method outperforms SOTA as it is, it can be further improved on real images to allow for complex texture patterns such as plaid, and specularities. We leave that for future work, for example, via improvement of projection to latent space which is an active research area in general GAN synthesis. We do demonstrate the full capabilities of our optimization method through try-on results on generated images. On generated images, our method can indeed further transfer high frequency details such as geometric patterns and complex textures, thus once latent space projection will be solved it will automatically apply to improve real images even further. 

Key contributions of this paper:
\begin{itemize}
    \item Photorealistic blend synthesis of person and garment (try-on) via StyleGAN interpolation algorithm using unpaired data.
    \item First method to allow for \textbf{body shape} deformation across the person and garment images.
    \item High quality synthesis of identity features and \textbf{body shape and skin color}. 
\end{itemize}

\section{Related Work} 

Virtual Try-On (often abbreviated as VITON and VTON) has seen tremendous progress in the recent years. Given a pair of images (person, garment), the original VITON method~\cite{VITON}  synthesizes a coarse try-on result, later refined, and warped with thin plate splines over shape context matching \cite{shape_matching}. CP-VTON~ \cite{CPVITON} adds geometric alignment to improve the details of the transferred garment. \citet{geometric} incorporates adversarial loss in \citet{CPVITON} to further improve image quality. PIVTONS~\cite{pivton} applies a similar concept on shoes rather than tops and shirts. \citet{multipose} extends \citet{VITON} to synthesize try on in various body poses. Further, GarmentGAN~\cite{GarmentGAN} separates shape and appearance to two generative adversarial networks. SieveNet~\cite{SieveNet} introduces a duelling triplet loss to refine details. ACGPN~\cite{ACGPN} aims to preserve the target person's identity in addition to the transferred clothes details, by accounting for semantic layout. SwapNet~\cite{swapnet} first warps and then applies texture to transfer full outfits, rather than individual garments. M2E Try-On Net~\cite{m2e} consists of three stages: a pose-alignment network, a texture refinement network, and a fitting network. \citet{m2e} uses an unpaired/paired training scheme that leverages images of the same person wearing the same clothes in different poses to improve textures and details in the synthesized try-on image. \citet{liquid_warping} proposes an Attentional Liquid Warping GAN with Attentional Liquid Warping Block (AttLWB). This method synthesizes high resolution results, but requires paired data of the same person wearing multiple different outfits. In contrast, our method does not require paired data.

\citet{ADGAN} and \citet{Zalando} incorporate learnings from StyleGAN \cite{StyleGAN} into try-on. ADGAN~\cite{ADGAN} conditioned the model on body pose, person identity, and multiple garments, where a separate latent code is generated to each of those components, and then combined into a single result by borrowing the needed parts from each image. This typically results in good transfer of uniform colors and textures but fails to synthesize the correct garment shape and texture details. \citet{Zalando} similarly conditions on pose and clothing items, but not for person's identity.  

A key assumption of all above methods is availability of large \textit{paired} training data, e.g.,  photographs of same person in various body poses wearing the same garment, or photograph of a person wearing a garment paired with separate garment images. Paired training data provides a ground-truth and a simpler design of losses. It is, however, a big limitation that tampers with quality and photo-realism of the results, since such paired data hard to obtain in large quantities required to train deep networks. 

O-VITON~\cite{OVITON} works with unpaired training data. It contains three stages: shape generation network, appearance generation network, both based on pix2pixHD \cite{pix2pixhd}, and an appearance refinement step. The shape and appearance generation network outputs are compared with the input image and segmentation in the loss function, the appearance refinement step is applied to each garment separately. The separation to three stages is what allows to work with unpaired data. Our algorithm, too, works with unpaired data, but with the key difference of doing all three stages in a \textit{single} optimization within the \citet{StyleGAN2} architecture. By eliminating the need for three separate steps, as well as our StyleGAN2 conditioning, we enable higher photo-realism. Furthermore, our method is able to accurately transfer garments across different body shapes, whereas \citet{OVITON} only shows results for a narrow range of body shapes. Since there is no public code or data for \citet{OVITON}, we cannot do a direct comparison. However, our high resolution results and focus on body shape differentiate our method from \citet{OVITON}. \citet{human_appearance} uses 3D pose and body shape information, barycentric procedures, and deep learning to synthesize try-on images in a wide variety of poses, however, their results are not photorealistic. 

Conditional GAN networks \cite{conditional_gans,cGANs} and GAN editing and inversion methods \cite{interpolation, NoiseInjection,Edo,I2I,GANWarp,Image2Stylegan,encoding_in_style,in_domain} are also related to our method. \citet{NoiseInjection} uses a grid structure to inject noise into a GAN to achieve spatial disentanglement on a grid, and then edit the image.  \citet{Edo} further accounts for spatial semantics by using K-means clustering to calculate spatial overlap between the StyleGAN activation tensors and semantic regions of an image. \citet{Edo} then uses a greedy algorithm to find interpolation coefficients that maximize changes within a semantic region of interest while minimizing changes outside of the region of interest. This method is a baseline for our proposed algorithm. All of these GAN editing algorithms are not focused on apparel try-on, but mostly on face photos. Running those for try on, as is, does not produce good results as we show in the ablation part. 
\input{03_methods}

\input{04_results}
\input{05_discussion}
\begin{acks}
We thank Edo Collins, Hao Peng, Jiaming Liu,  Daniel Bauman, and Blake Farmer for their support of this work.
\end{acks}
{\small
\bibliographystyle{ACM-Reference-Format}
\bibliography{SIGGRAPH}
}

\end{document}

%% file: 03_methods.tex
\section{Method}
\begin{figure}
         \centering \includegraphics[width=.4\textwidth]{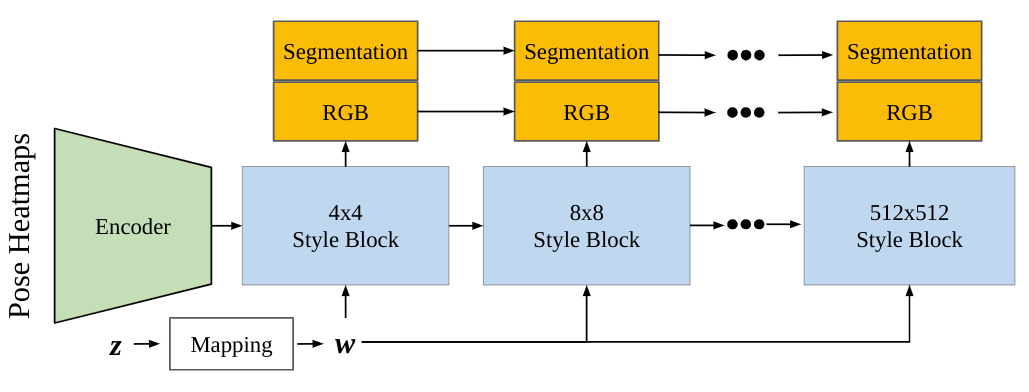}
         \caption{We trained a pose conditioned StyleGAN2 network, that outputs both an RGB image as well as clothing segmentation of the image in each layer. Pose heatmaps are encoded and inputted into the first style block in StyleGAN2 instead of a constant input.}
         \label{fig:pose-network}
\end{figure}

\begin{figure}
    \centering
    \includegraphics[width=0.4\textwidth]{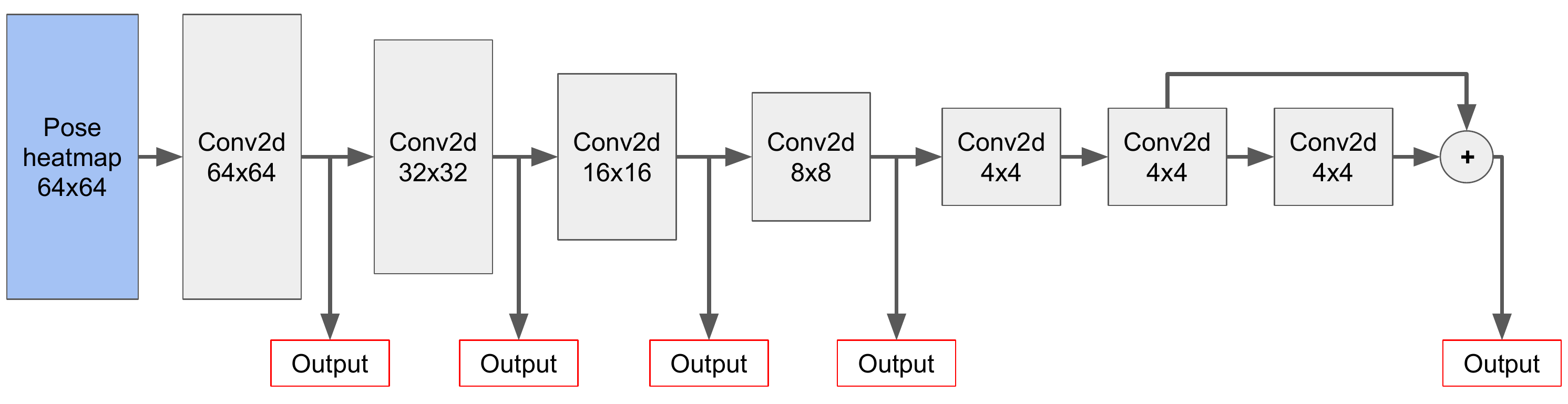}
    \caption{Pose encoder architecture for the pose-conditioned StyleGAN2. Our multi-resolution encoder has an output for each resolution from $4\times 4$ to $64\times 64$. The final $4\times 4$ output of the encoder is multiplied by $\frac{1}{\sqrt{2}}$ and is the input to our StyleGAN2 generator. The other resolution outputs are concatenated with the up-sampled input to each style block, beginning with the $8\times 8$ style block. }
    \label{fig:encoder}
\end{figure}

\begin{figure*}
         \centering
         \includegraphics[width=\textwidth]{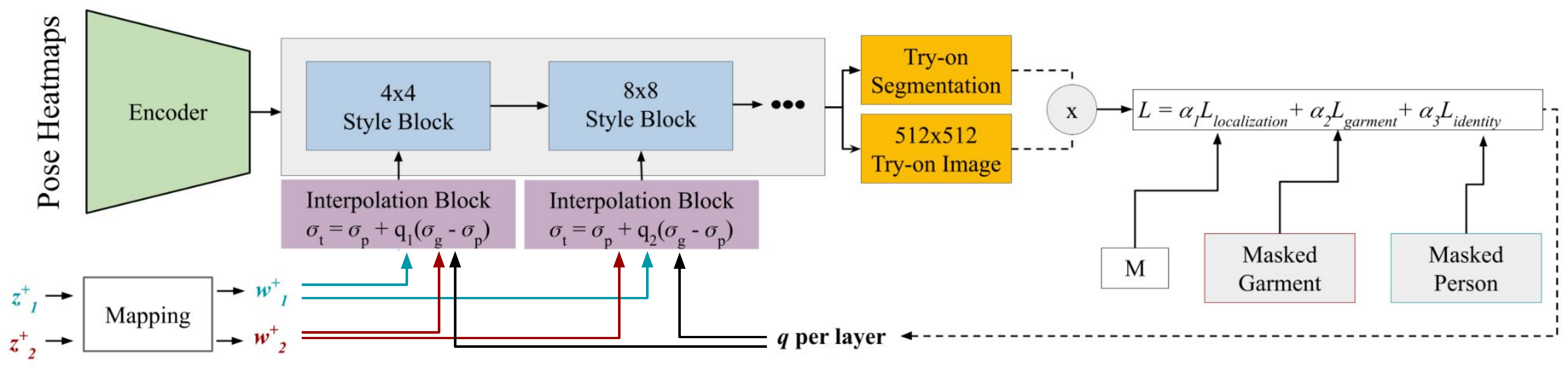}
         \caption{ The try-on optimization setup illustrated here takes two latent codes $z_1^{+}$ and $z_2^{+}$ (representing two input images) and an encoded pose heatmap as input into a pose-conditioned StyleGAN2 generator (gray). The generator produces the try-on image and its corresponding segmentation by interpolating between the latent codes using the interpolation-coefficients $q$. By minimizing the loss function over the space of interpolation coefficients per layer, we are able to transfer garment(s) $g$ from a garment image $I_g$, to the person image $I_p$.}
         \label{fig:interpolation_network}
\end{figure*}

In this section, we describe our TryOnGAN optimization algorithm for garment transfer. Given a pair of images generated by StyleGAN2, we show how to optimally interpolate between the generated images to accomplish try-on. We also describe how to use the network for any real image, via projection of the image to our latent space, and then running TryOnGAN. 

\subsection{Problem Formulation} 

Given an image $I^p$ of a person $p$ in some outfit, and an image $I^g$ of a different person in a garment $g$, we aim to create a photo-realistic synthesis of the person $p$ in garment $g$. 

The first step of our algorithm is to train a pose conditioned StyleGAN2, which can generate a photorealistic image of a person in some outfit given a 2D pose skeleton. We train our model to output RGB images as well as the garment and person segmentation in the image.  Given a trained model, the second step is to optimize for interpolation coefficients at each layer to get the desired try-on result image $I^t$ where person $p$ will appear in garment $g$. 

\subsection{Pose-conditioned StyleGAN2}
Generative Adversarial Networks (GANs)~\cite{gans} have been shown to synthesize impressive images from latent codes. StyleGAN and StyleGAN2 \cite{StyleGAN,StyleGAN2} in particular demonstrated state-of-the-art photo-realism on face images. The idea to combine progressive growing~\cite{progressive_growing} and adaptive instance normalization
(AdaIN) \cite{AdaIN, artistic_style,feature_wise,arbitrary_neural_artistic} with a novel mapping network between the latent space, $Z$, and an intermediate latent space, $W$, encouraged disentanglement of the latent space. Transforming intermediate latent vector $w\in{W}$ into style vectors $s$ further allowed different styles at different resolutions. Furthermore, recently developed StyleGAN inversion methods enable the projection of real images into the extended StyleGAN latent spaces, $Z^+$ and $W^+$ \citet{Image2Stylegan}. Motivated by those advances we choose StyleGAN2 as our base architecture.

\begin{figure}
    \centering
    \includegraphics[width=0.35\textwidth]{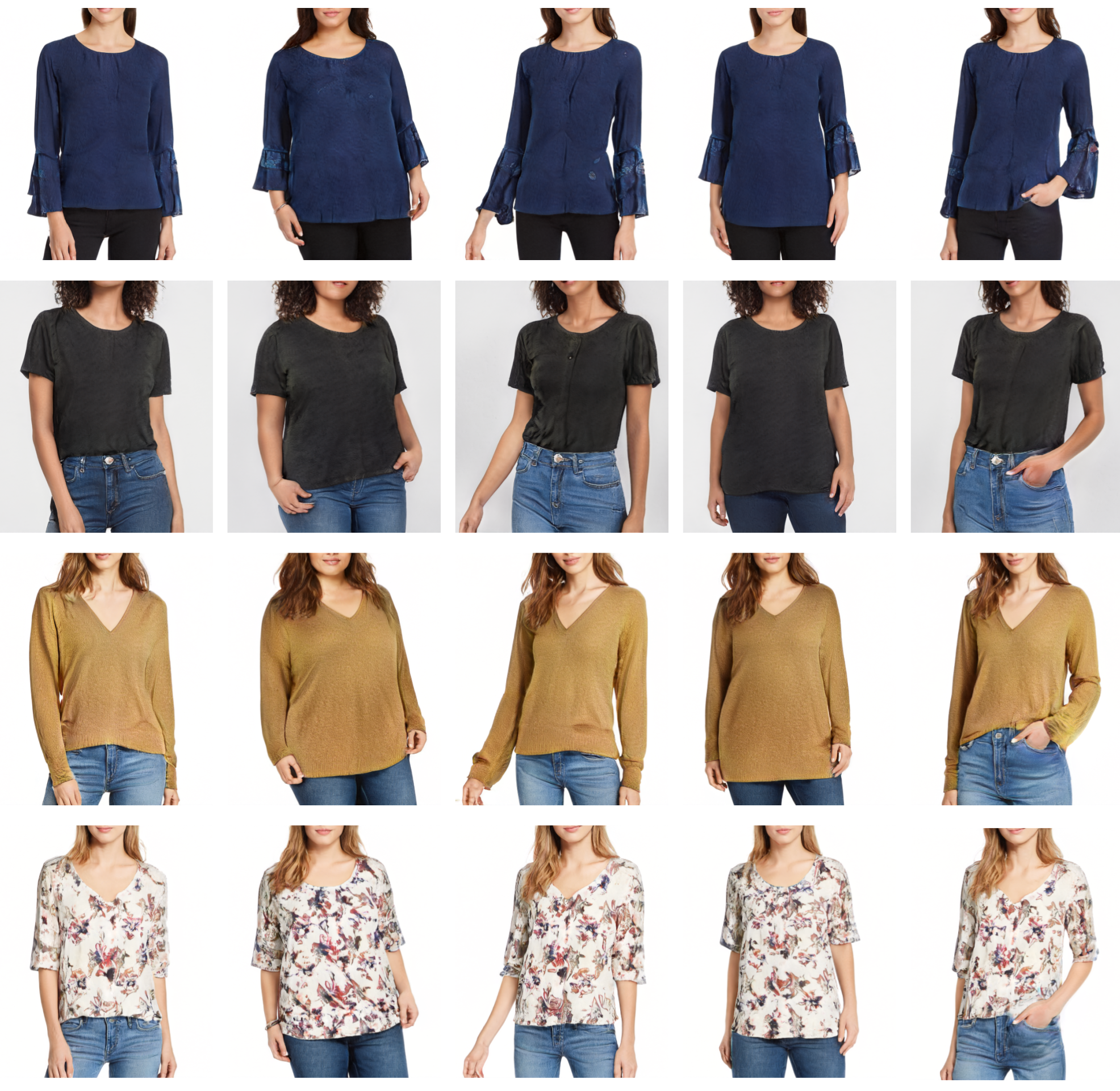}
    \caption{Our method can synthesize the \textit{same style shirt} for varied poses and body shapes by fixing the style vector. We present several different styles in multiple poses. In this figure, each row is a fixed style, and each column in a fixed pose and body shape. }
    \label{fig:poses}
\end{figure}

\begin{figure}
    \centering
    \includegraphics[width=0.3\textwidth]{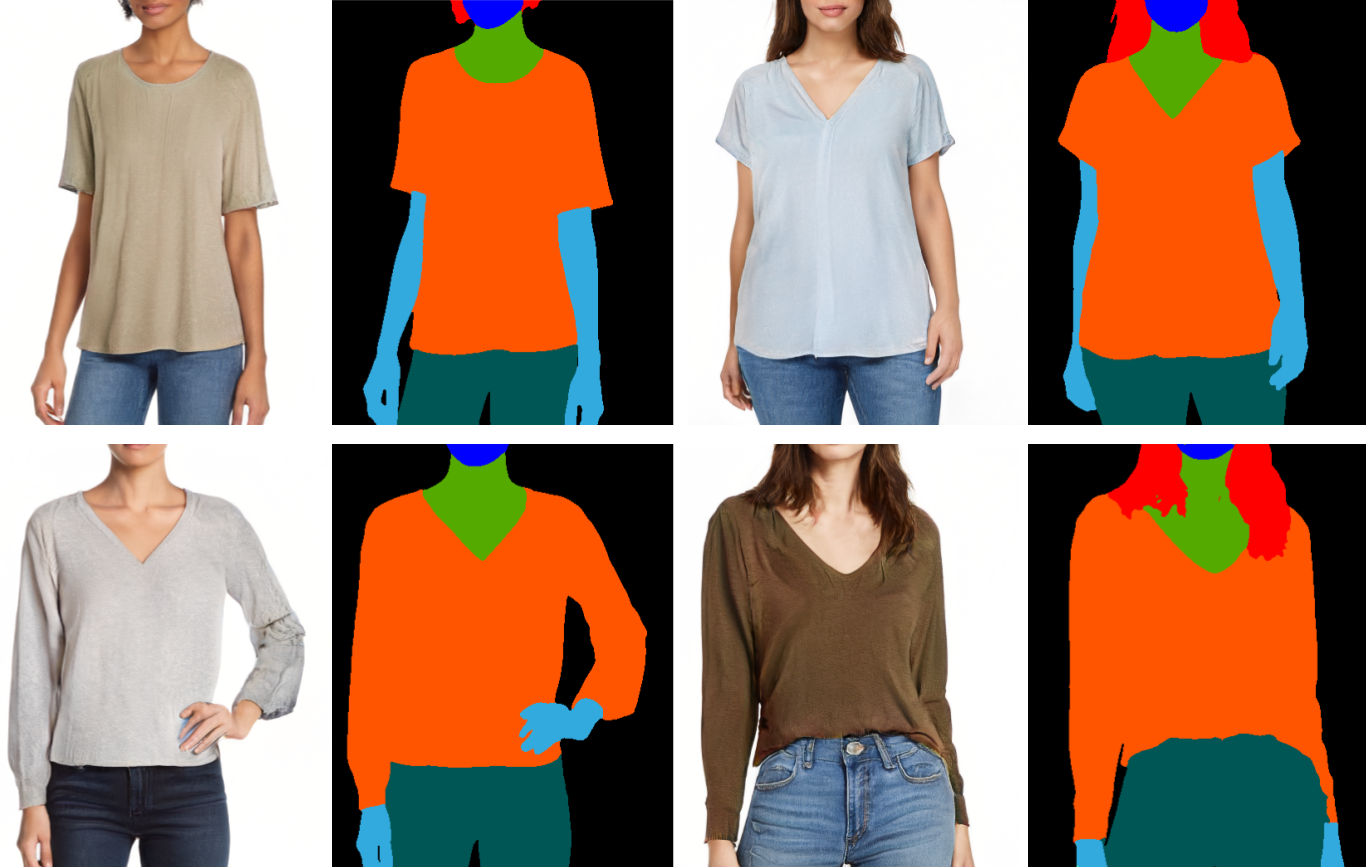}
    \caption{Example images and segmentations generated by our TryOnGAN.}
    \label{fig:segmentation}
\end{figure}


We train a StyleGAN2 model on fashion images, with three key modifications (Figure \ref{fig:pose-network}). We explain the details and justifications of each modification below.

\subsubsection{Pose-conditioning} 

Without pose conditioning, the latent space entangles pose and style, thus changing pose during try-on. To prevent entanglement, we condition our StyleGAN2 network on pose. We replace the constant input to the beginning of the generator with an encoder (Figure \ref{fig:encoder}) that takes as input a $64\times 64$ resolution pose representation. In our experiments, we use PoseNet \cite{tensorflow} to obtain 17 pose keypoints and keypoint confidence scores for each image in our dataset. We create a 17-channel heatmap, where each channel corresponds to a particular keypoint (channels corresponding to cropped-out keypoints are set to zero). Each heatmap channel is created by generating a small gaussian (sigma = 0.02 * heatmap width) centered at the location of the keypoint and weighted by the keypoint confidence score. These $64\times 64$ resolution pose heatmaps are the pose representation input to our encoder.
See Figure \ref{fig:poses} for pose-conditioned images generated by our model.

\subsubsection{Segmentation Branch} Our model outputs segmentations as an auxiliary task to improve disentanglement and to remove dependencies on existing segmentation models during optimization. Our segmentation branch follows the StyleGAN2 RGB branch architecture. See Figure \ref{fig:segmentation} for examples of segmentations and RGB images generated by our model.

\subsubsection{Discriminators} We train using two discriminators, one for pose and one for segmentation. The pose discriminator receives as input either a real RGB image/pose heatmap pair or a conditional pose heatmap with the corresponding generated RGB image. The segmentation discriminator receives real/generated RGB image/segmentation pairs. The two discriminators are weighted equally during training. To prevent overfitting of pose to style, we use the following data augmentations on the pose input to the discriminator: 1) add gaussian noise to the normalized keypoint locations before creating the heatmap and 2) drop keypoints with probability less than 0.4.

Unlike other methods \cite{CPVITON,ADGAN}, our method can be trained on un-paired data. At training time, the only requirements are pre-computing segmentation and pose for the dataset. These are  obtained using existing methods \cite{tensorflow,Graphonomy}.


\subsection{Try-On Optimization}

Given the trained model, we can generate a variety of images (along with the corresponding segmentations) within the latent space of the network with the desired 2D pose. Conversely, given an input pair of images, we can "project" the images to the latent space of the generator by running an optimizer to compute the latent codes that minimizes the perceptual distance between the input image and the image from the generator.  Linear combinations of these latent codes will produce images that combine various characteristics of the pair of input images. The desired try-on image where the garment from the second image is transferred to the person from the first image lies somewhere within this space of combinations. Let us denote by  $\sigma _ \textit{p}$ and $\sigma _ \textit{g}$ the style scaling coefficients per layer for person and garment images respectively. Interpolation between the style vectors can be expressed as: 
\begin{equation} \label{eq:interpolation}
    \sigma _t = \sigma_p + Q(\sigma_g - \sigma_p)
\end{equation}
where \textit{Q} is a positive semi-definite diagonal matrix. The elements along the diagonal form a query vector, \textit{q} $\in$ [0,1]. Generating the try-on image can be accomplished by recovering the correct interpolation coefficients $q$.  \citet{Edo} proposes a greedy algorithm to choose binary query vectors that maximize changes within the region of interest while minimizing changes outside of the region of interest.  

We also optimize over the query vectors but instead of greedy search for a set of coefficients using a fixed budget for every layer as in \citet{Edo}, we propose an optimization-based approach that allows for more flexibility in the choice of query vectors (see Figure \ref{fig:editing_comparison} for ablation study results). Furthermore, \citet{Edo} requires \textbf{manual} clustering (labor intensive process) to define semantic regions while our method is fully automatic. Our optimization loss terms are tuned to the try-on problem of preserving the identity of the person while switching the garment of interest. The loss functions in our optimization guide our method to learn continuous query vectors that enable more localized semantic edits.

Let \textit{S$^p$}, \textit{S$^g$}, and \textit{S$^t$} be the segmentation labels corresponding to $I^p$, $I^g$, and $I^t$. We pre-compute \textit{S$^p$}, \textit{S$^g$} since $I^p$, $I^g$ do not change during the optimization. We use the $512\times 512$ segmentation output of our network for \textit{S$^t$} since $I^t$ is updated each optimization iteration (segmentation weights are frozen and do not change with respect to the optimization loss). Figure~\ref{fig:interpolation_network} presents the flow of our algorithm. We modify our pose-conditioned StyleGAN2 to take in intermediate latent codes, $z^{+}_1$ and $z^{+}_2$, from the extended $Z^+$ space for both the person and garment images. \citet{Image2Stylegan} showed that sampling a latent vector per StyleGAN layer, rather than one latent vector for all layers, improves results. We use their notation, z+ and w+, for these extended latent vectors. (Eq. \ref{eq:interpolation}) occurs in every style block and the generator outputs the try-on image and segmentation. The generator is conditioned on the pose of $I^p$, such that $I^g$ is re-posed to the pose of $I^p$. The outputs are combined to calculate the loss terms which are optimized over the space of interpolation-coefficients $q$ until convergence, where the dimension of $q$ is 16 (number of layers) x number of channels per each corresponding layer. Our loss function is defined as follows: 

\begin{equation}\label{interpolation}
    L =  \alpha_1 L_{\textit{localization}} + \alpha_2 L_{\textit{garment}} + \alpha_3 L_{\textit{identity}} 
\end{equation}
where the $\alpha$s are weights applied to the loss terms and are hyper-parameters for our method. At each iteration, the $q$ vector values are clipped to $[0,1]$ using a sigmoid after applying the updates. Each of the loss terms is described below.


%



\subsubsection{Editing-localization Loss}
The editing-localization loss term encourages the network to only interpolate styles within the region of interest. The region of interest (e.g. shirt or pants) is chosen at test time. Similar to \citet{Edo}, we define a term, \textit{M}, that measures spatial overlap between the semantic regions in the image and the activation tensors, $A_{\text{NxCxHxW}}$, where N is the number of images, C is the number of channels, and H, W are the image dimensions. \citet{Edo} uses k-means on the activation tensors to manually assign semantic cluster memberships to the activation tensors. Instead, we use the segmentation outputs from our network to define semantic cluster memberships. The segmentations are converted to binary cluster membership heatmaps, \textit{U} $\in  \{0,1\}^{\text{NxKxHxW}}$, where K is the number of segments. For each layer, the heatmaps are downsampled to the correct resolution and the activation tensors are normalized per channel by subtracting the mean of each channel and dividing by the standard deviation of the channel. \textit{M} is then computed as:

\begin{align}\label{M}
    M_{k\times c} = \frac{1}{NHW} \sum_{n,h,w} A^2 \odot U
\end{align}

We calculate M per layer for the person and garment images. We denote these $k\times c$ matrices as $M^p_{k\times c}$ and $M^g_{k\times c}$. For a segment of interest, $i$, we calculate the least relevant activation channels by subtracting the $i$th row of each M matrix from the k rows in that matrix. We take the max over each channel (each column) to get a c-dimensional vector. We perform an element-wise max over the c-dimensional vectors corresponding to person and garment. 

\begin{align} \label{relative_M}
    &M^{p'}_{k\times c} = M^p_{j,:} - M^p_{i,:} \text{ for } j=0,\cdots,k-1  \\
    &M^{p'}_c = \max(M^{p'}_{k\times c}) \\
    &M^{g'}_{k\times c} = M^g_{j,:} - M^g_{i,:} \text{ for } j=0,\cdots,k-1 \\
    &M^{g'}_c = \max(M^{g'}_{k\times c}) \\
    &M^i_c = \max(M^{p'}_c, M^{g'}_c)
\end{align}
High values in \textit{$M^i_c$} represent the channels that correspond to segments other than $i$ in either image. Since we only want to change the segment of interest, \textit{i}, we want the interpolation coefficients for all other segments to be low. Therefore the editing-localization loss term is computed as: 

\begin{align}\label{M_loss}
    &L_{\textit{localization}} = \sum M^i_c \odot q_c 
\end{align}

\subsubsection{Garment Loss}
To transfer over the correct shape and texture of the garment of interest, we use VGG embeddings \cite{VGG,vgg_perceptual} to compute the perceptual distance between the garment areas of the two images. Given the segmentation labels $S^g$ and $S^t$ corresponding to the garment and try-on result images, we compute binary masks for the garment in both images. We apply the mask to the RGB images by element-wise multiplication, followed by blurring with a gaussian filter and downsampling to 256 x 256 before finally computing the garment loss $L_{\textit{garment}}$ as the perceptual distance between the two masked images. 

\begin{align}\label{garment_loss}
    L_{\textit{garment}} = d(I^{g}_{\textit{Garment Masked}}, I^{t}_{\textit{Garment Masked}})
\end{align}
where \textit{$d(\cdot,\cdot)$} measures the perceptual distance by calculating a weighted difference between VGG-16 features.


\subsubsection{Identity Loss}
The identity loss term guides the network to preserve the identity of the person \textit{p}. We use the hair and face regions of the images as a proxy for the identity of the person. Using the segmentation labels $S^p$ and $S^t$ corresponding to the person image $I^p$ and the try-on image $I^t$, we compute the identity loss following the same procedure as the garment loss above. 

\begin{align}\label{identity_loss}
    L_{\textit{identity}} = d(I^{p}_{\textit{Identity Masked}}, I^{t}_{\textit{Identity Masked}})
\end{align}
%


%


\subsubsection{Projection}\label{section:methods-optimization-projection}
To run our algorithm on real images, we first project the real images into an extended latent space, \textit{Z+} \cite{Image2Stylegan}. We use an optimization to learn a latent vector, \textit{z}, per layer that results in a final image that best captures the identity and garment details of the original image. The optimization uses a perceptual loss \cite{vgg_perceptual} to find the optimal latent vectors. We project using our pose-conditioned network and condition on the pre-computed pose of the image being projected.

%% file: 04_results.tex
\begin{figure}
    \centering
    \includegraphics[width=0.3\textwidth]{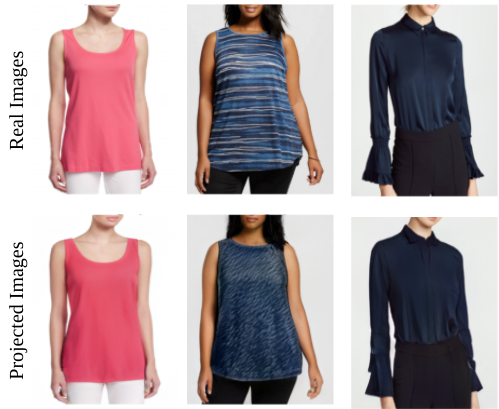}
    \caption{Typical projection examples. It is useful to see the effect of projection on the quality of the garment representation, since it directly impacts the final try on result. Improving the projection is independent of our optimization algorithm and is part of future work.}
    \label{fig:projection}
\end{figure}

\begin{figure}
    \centering
    \includegraphics[width=0.45\textwidth]{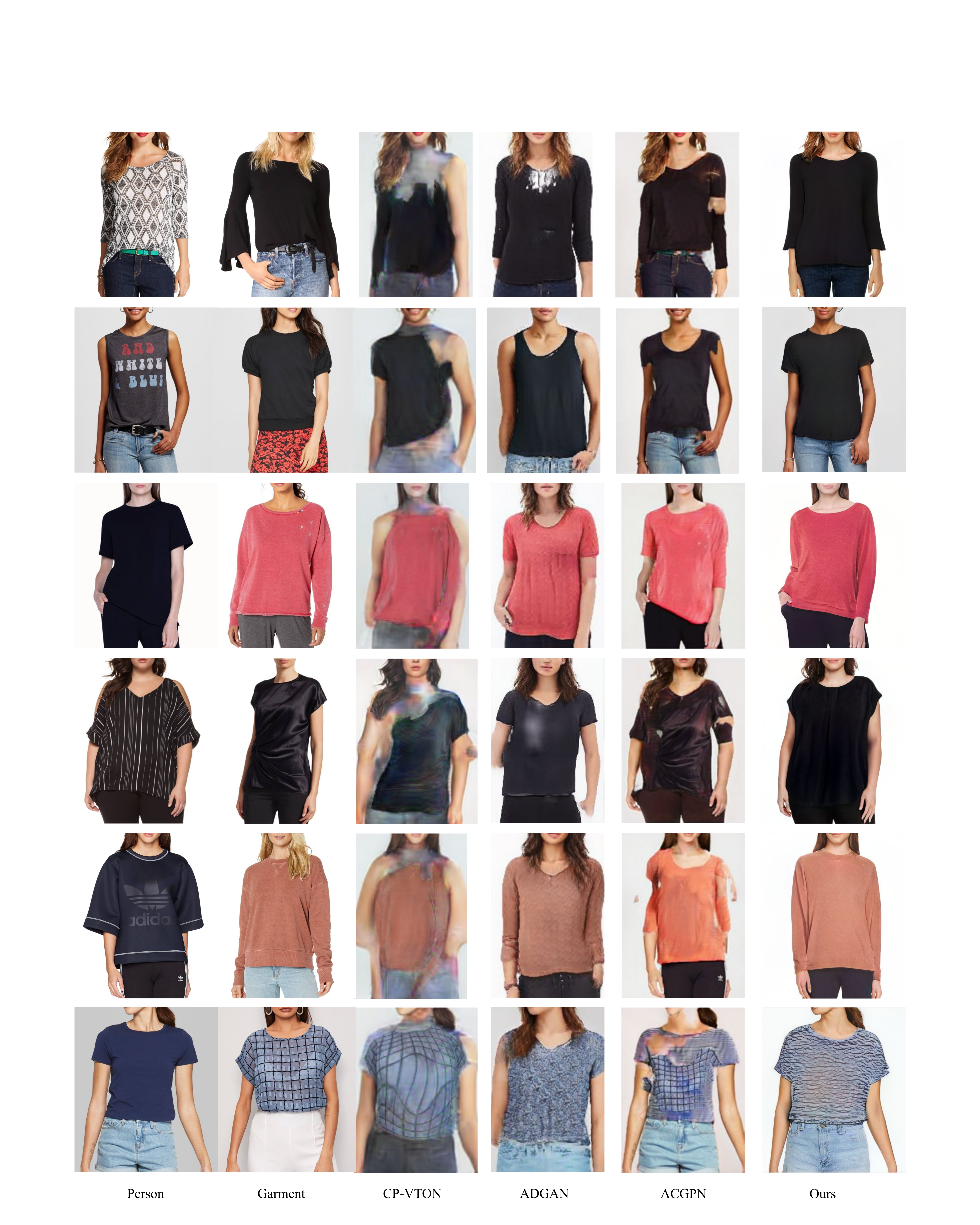}
    \caption{Qualitative comparison with \citet{CPVITON}, \citet{ADGAN}, and \citet{ACGPN} on real image try-on. Each row represents a different pair of inputs. Note the difference in garment quality, adjustment to difference in body shape, skin color, and pose. TryOnGAN outperforms the state of the art significantly.}
    \label{fig:qualitative_baselines}
\end{figure}

\section{Experiments}

In this section, we provide implementation details, comparison to related works, ablation studies, and results from our method on diverse examples. 

\subsection{Dataset} We collected a dataset of people wearing various outfits, and partition it into a training set of 104K images, and a test set of 1600 images. All results are shown on the test set images. The resolution of all the images is  512x512 (images cropped in figures for ease of visualization). The dataset includes people of different body shapes, skin color, height, and weight. We focus in this work on females only, and perform try on for tops and pants. We show additional results from our method on the public Street2Shop dataset \cite{street2shop}. 

\subsection{Implementation details} Our conditional StyleGAN2 network was implemented in TensorFlow. We trained it for 25 million iterations, on 8 Tesla v100 GPUs, for 12 days. Once the network was trained, we performed a hyperparameter search for the optimization loss weights in Eq. \ref{interpolation}. For generated images, we used $\alpha_1=0.01$, $\alpha_2=1$, and $\alpha_3=0.2$. For real images, we used $\alpha_1=0.01$, $\alpha_2=1$, and $\alpha_3=1.0$. We used the Adam optimizer for the optimization method. The try-on optimization method was run for 2,000 iterations per pair for both real and generated images. The real images were first projected into the StyleGAN2 latent space by running the projection optimization for 2,000 iterations. The average runtime of our try-on optimization method is 224.86s with a std of 0.38s per pair of images. The average runtime of our projection optimization is 227.77s with a std of 4.29s per pair of images.

\subsection{Comparison to Virtual Try-On Methods on Real Images}

We compare quantitatively and qualitatively to two state-of-the-art virtual try-on methods with code available: ADGAN \cite{ADGAN} and CP-VTON \cite{CPVITON}. We use the available pre-trained weights for ADGAN and CP-VTON since they require paired data to train, which we don't have. We additionally include qualitative comparisons with a recent baseline, ACGPN \cite{ACGPN}, and use the available pre-trained weights.

\subsubsection{Image projection} The first step is projecting the garment and person image into the latent space. The quality of projection impacts the final try-on images. In Figure~\ref{fig:projection} we show examples of real images and the corresponding projected images. We use the standard StyleGAN2 projection method extended to the Z+ space as described in \ref{section:methods-optimization-projection}. Note that the projection used is independent of our optimization method (see Figure \ref{fig:shirts} for interpolation without projection). Therefore improving projection as future work would continue to improve our final try-on images, however, even as it is now it outperforms SOTA. 

\begin{figure}
    \centering
    \includegraphics[width=0.45\textwidth]{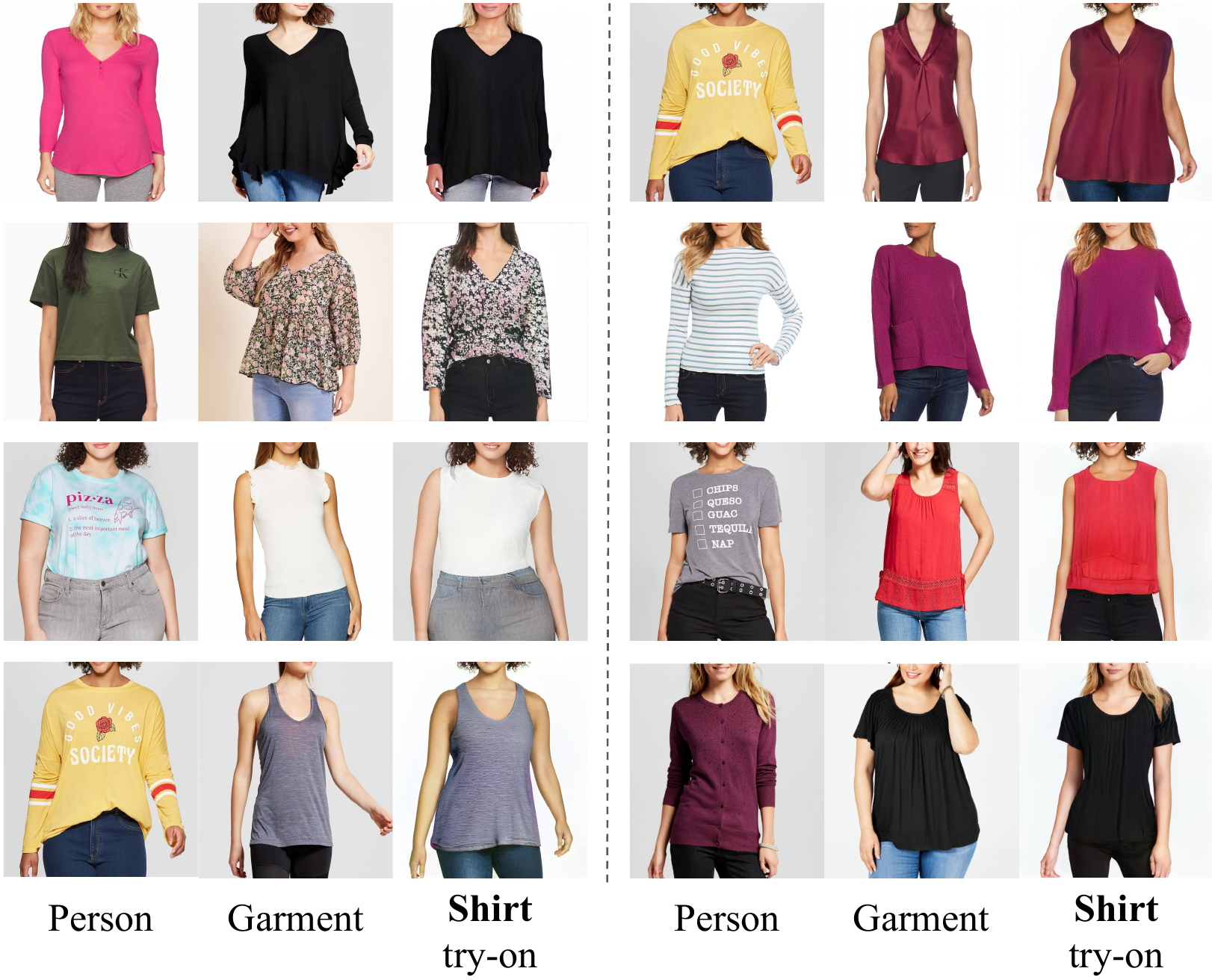}
    \caption{Results from our method for \textbf{shirt} try-on on \textbf{real} images. Note how try-on works well with different body shapes, and adjusts to the new poses. Some details are missing from the garment due to artifacts in projection, however the overall shape is well preserved.}
    \label{fig:shirt_real}
\end{figure}


\subsubsection{Qualitative Evaluation} Figure \ref{fig:qualitative_baselines} compares virtual try-on results produced by our TryOnGAN method with those produced by the baselines for various shirt and body types. Our method is able to synthesize the correct shape of the shirt and preserve high frequency details. In cases where the target shirt has a shorter sleeve while the person is wearing a longer sleeve, TryOnGAN is able to accurately synthesize the arms and preserve body shape and skin color. In comparison, ADGAN and ACGPN are unable to synthesize the correct shape of the try-on shirt (e.g. neckline and sleeve length) and are unable to synthesize correct body shape. ADGAN synthesizes the correct color of the shirt and coarse identity information, but is unable to preserve details for both the garment and identity. There are also several artifacts that prevent the try-on image from being photo-realistic. CP-VTON copies the hair and face of the person to the try-on image, but is unable to accurately synthesize the person's body. CP-VTON preserves the color of the garment, however the final try-on is typically blurry. Figure \ref{fig:shirt_real} shows additional results from our method on real images.
 
\begin{table}[tb]
\centering
\caption{Quantitative measure of our method and the baselines. We use two metrics to compare the methods and types of images: FID to evaluate photorealism and ES (Embedding similarity) to evaluate quality of try-on or how similar is the result to the input in the garment part. We also include user study results, which indicate the percentage of participants who preferred results from each method. All results are computed on try-on results for real images. }
\begin{tabular}{ |c|c|c|c|  }
 \hline
 Model & FID $\downarrow$ & ES $\uparrow$ & User Study\\ 
 \hline 
 ADGAN \cite{ADGAN} &  66.82  & 0.22 & 31.3\%  \\
 CP-VTON \cite{CPVITON} & 87.0   &  0.27 & 6.1\% \\
 Our Try-on on Real  & \textbf{32.21} & \textbf{0.32} & \textbf{62.6\%}  \\ 
 Real Images & 11.83 & N/A & N/A  \\
 \hline
\end{tabular}
\label{table:quantitative}
\end{table}

\subsubsection{Quantitative Evaluation}
We evaluate the results using quantitative measures, Fr\'{e}chet Inception Distance (FID) \cite{FID}, embedding similarity (ES) score \cite{embedding}, and a perceptual user study.  Table \ref{table:quantitative} shows the FID, embedding similarity, and user study results. The FID and ES experiments were run over 800 images for each algorithm. We can see that our method outperforms others on FID score, which represents photorealism. For calibration, we have also calculated FID scores for a set of real images. The embedding similarity score measures the distance between embeddings of the original garment and the garment in the try-on image. Our method has the highest similarity which reflects our method's ability to preserve the shape and details of the try-on garment. We also ran a user study with 41 participants (20 from Amazon Mechanical Turk and 21 random participants from an institution mailing list). Each participant chose the best result between all possible pairs (ADGAN/CP-VTON, CP-VTON/Ours, ADGAN/Ours) for the six virtual try-on pairs in Figure \ref{fig:qualitative_baselines}. The order of pairs and order within pairs were randomized. We also included a few repeat questions (results only included once) to ensure participants understood the task and chose consistently. The results show that across the 41 participants, our method was preferred twice as often as ADGAN and preferred about ten times more often than CP-VTON.

\begin{figure}
    \centering
    \includegraphics[width=0.4\textwidth]{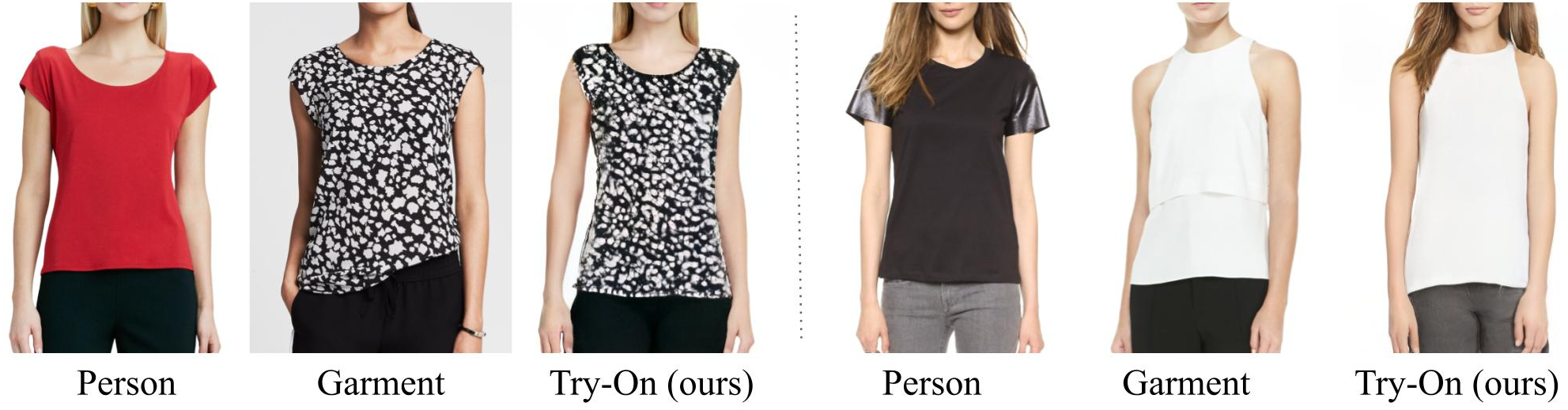}
    \caption{Try-on results by our method on real images from the Street2Shop dataset \cite{street2shop}. Our method is able to generalize to a new dataset without retraining.}
    \label{fig:street2shop}
\end{figure}

\begin{figure*}
    \centering
    \includegraphics[width=0.75\textwidth]{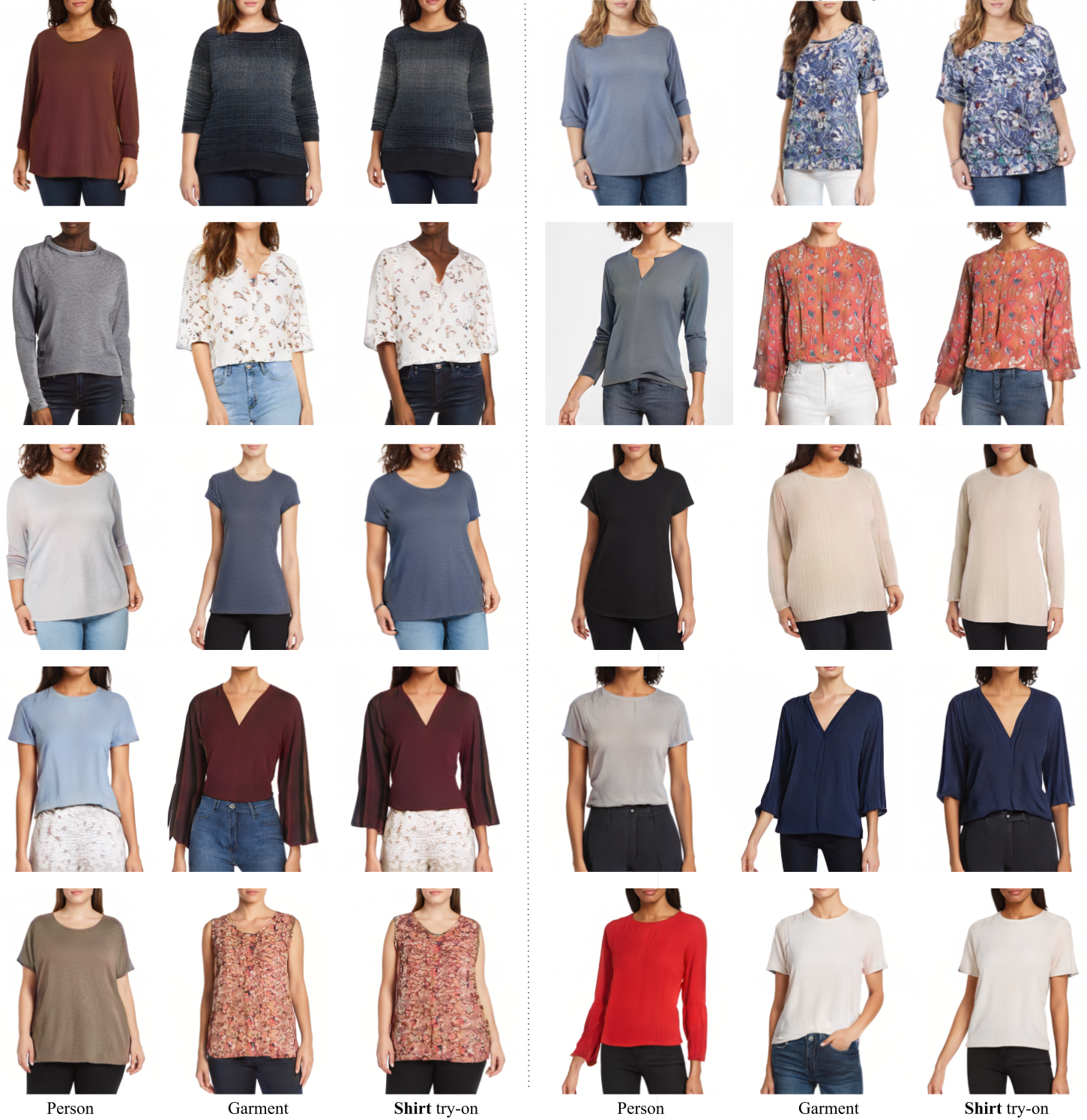}
    \caption{Results from our method for ten \textbf{shirt} try-on examples on generated images. Note how try-on works well with different body shapes, and adjusts to the new poses. Our method is able to transfer complex garment patterns and textures. \textbf{Zoom in for details.} }
    \label{fig:shirts}
\end{figure*}

\begin{figure}
    \centering
    \includegraphics[width=0.3\textwidth]{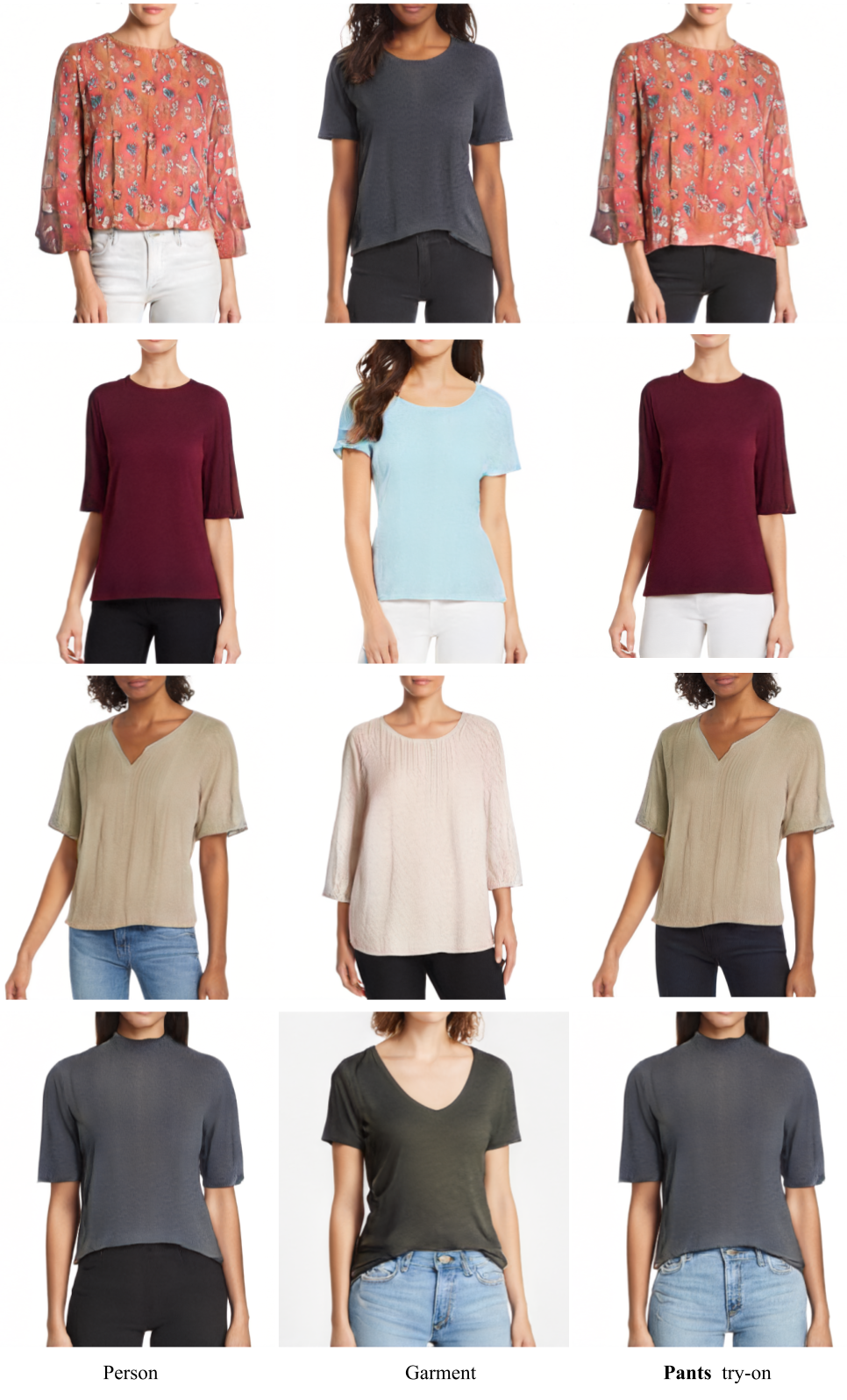}
    \caption{Results from our method for \textbf{pants} try-on on generated images. Note how try-on works well with different body shapes, and adjusts to the new poses. Our method can also synthesize garment details such as buttons and pockets that weren't in the original person image. Each row corresponds to a different try on result. Columns represent person, garment, and result.}
    \label{fig:pants}
\end{figure}






\subsubsection{Generalization to other datasets}
Figure \ref{fig:street2shop} shows results from our method on an additional dataset, Street2Shop \cite{street2shop}. Without retraining, our method is able to produce high resolution try-on images. 

\subsection{Try-On Results on Generated Images}

While TryOnGAN already improves on SOTA, we show that our optimization method has the capacity to improve even further, with no modifications, once projection (an active area of research) is solved.  Running our try-on optimization on generated images results in highly detailed images capable of capturing complex garment patterns and textures. 

\subsubsection{Results} Figure \ref{fig:teaser}, Figure \ref{fig:shirts}, and Figure \ref{fig:pants} show  try-on results produced by our method on generated images. Note the diversity of the people wearing the items, how the identity of the person is preserved even though the try-on output is synthesized from scratch, and the details on the transferred item (note neck lines, pattern, sleeve length, color). It is also worth noting the garment folds appearing on the new person, since that person might have different body shape, or pose. We present try on of both pants and shirts. Our method is also successful in preserving the skin-tone of the person in the input image though the garment image may contain a person with a different skin-tone. In the case of transferring a shorter sleeve length garment, our method synthesizes the arms appropriately though they were not visible in the input image.

\begin{figure}
    \centering
    \includegraphics[width=0.4\textwidth]{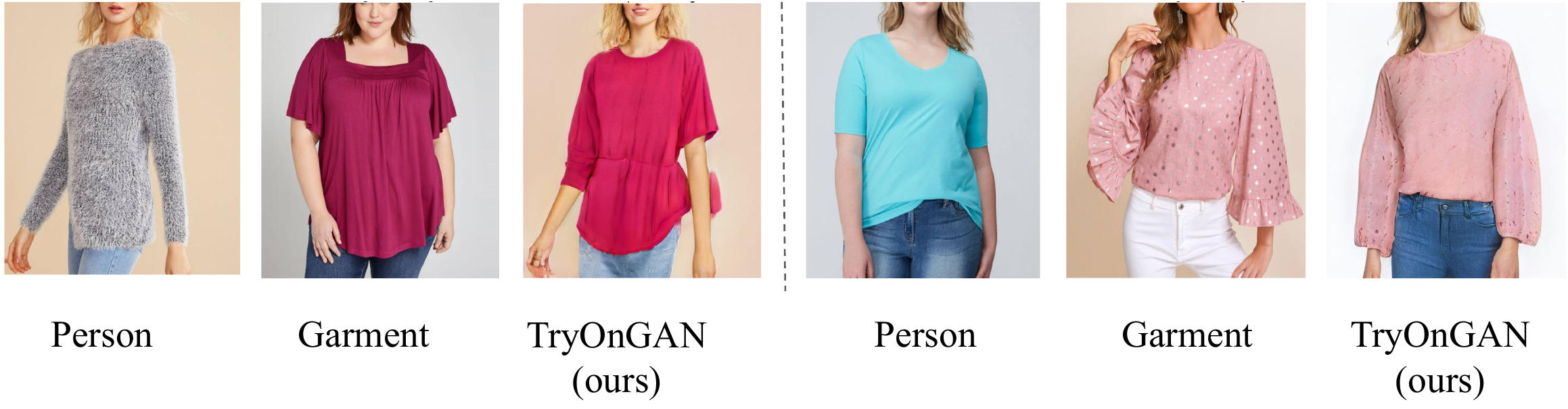}
    \caption{Failure cases for our method on real images. Our method typically fails when garment detail or pose wasn't represented well in the training dataset.}
    \label{fig:failure}
\end{figure}

\begin{figure}
    \centering
    \includegraphics[width=0.3\textwidth]{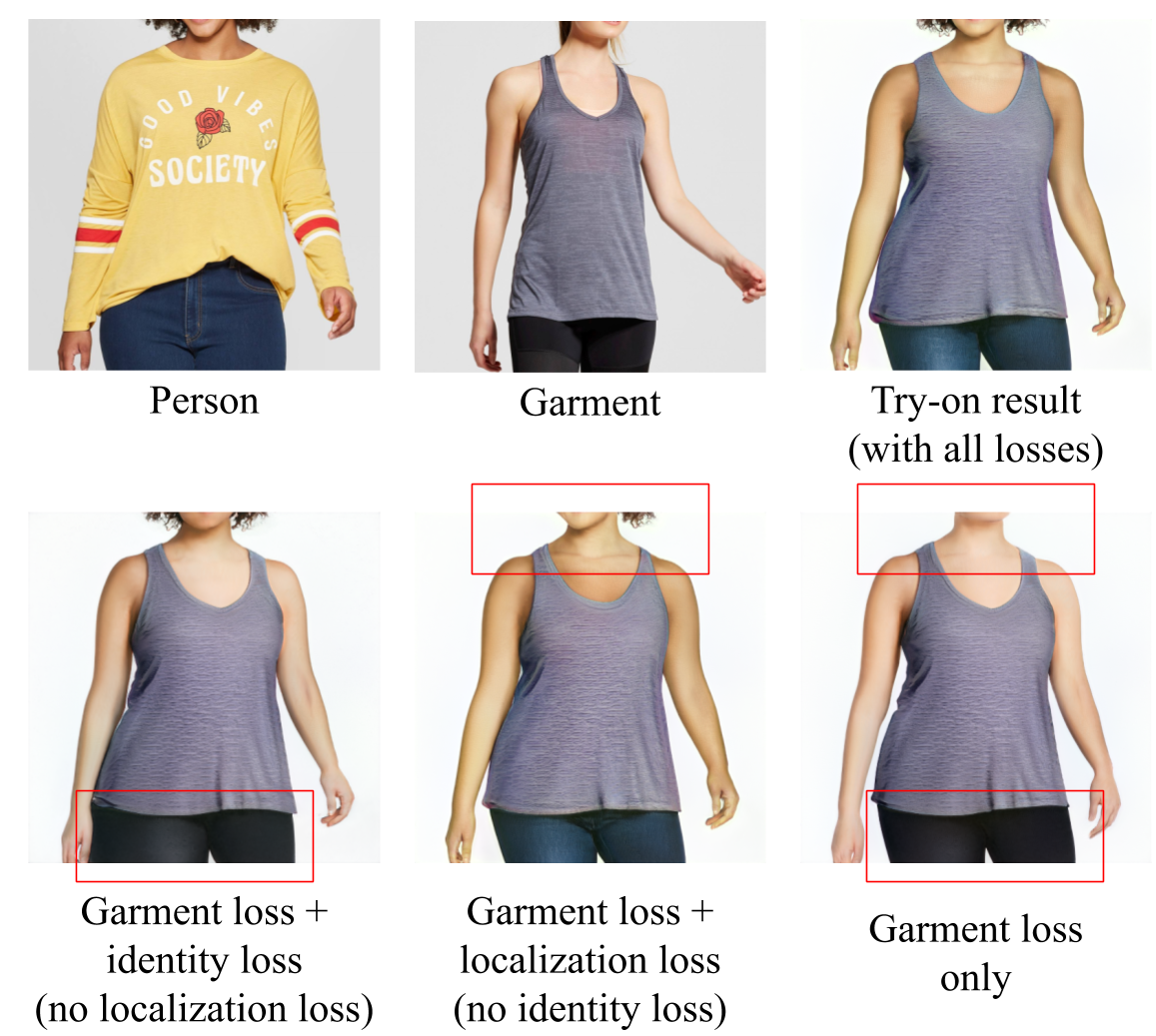}
    \caption{Ablation study showing the importance of each loss term in the optimization. The study is done on real images.}
    \label{fig:ablation_real}
\end{figure}


\subsection{Ablation study and failure cases}\label{sec:failure-cases}
Figure \ref{fig:failure} shows examples of when our TryOnGAN method fails to correctly synthesize try-on images. Rare poses (not well represented in the data) or rare garment details cause the appearance of the garment to change when transferred to the target pose. We suspect that the results for those would improve with better representation of diverse garments and poses in the training dataset and subsequently in the latent space. Similarly, as discussed above, projection of real images to latent space has artifacts which affect the try-on result. Once projection is improved, our method will be able to generate true to source results. 

\begin{figure}
    \centering
    \includegraphics[width=0.35\textwidth]{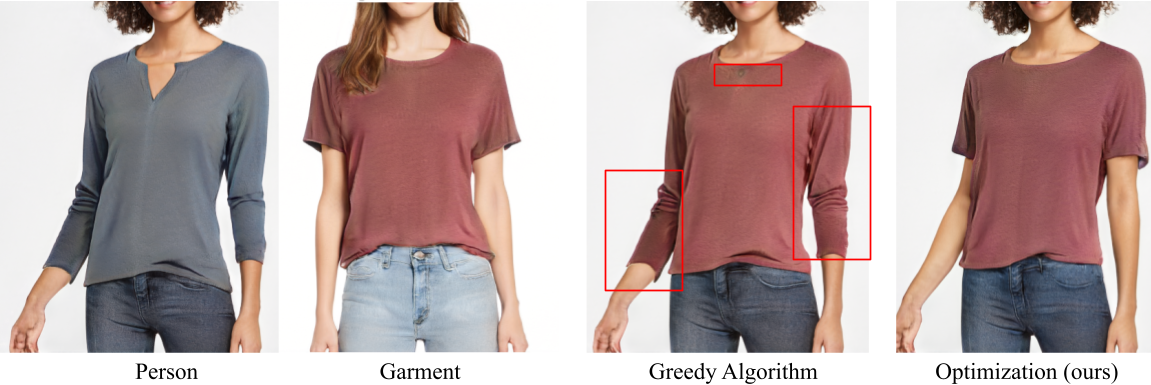}
    \caption{Ablation study: we compare greedy search for interpolation coefficients as in \citet{Edo} to our optimization approach on generated image try-on. We observe that details like sleeve length and pattern are preserved much better with the per layer optimization approach. Note that we do not compare directly to \citet{Edo} since we also modified the StyleGAN architecture to include segmentation and condition on pose. The red boxes highlight incorrect sleeve-length and artifacts generated by the greedy search method. }
    \label{fig:editing_comparison}
\end{figure}

Figure \ref{fig:ablation_real} shows how our result changes with changes to the loss function. We run this ablation study on real images and demonstrate that each loss term is necessary for a photo-realistic result that preserves garment characteristics and person identity. The localization loss prevents the optimization from editing semantic regions outside of the garment of interest. The identity loss preserves the face and hair. The garment loss transfers the shape, color, and texture of the garment.

Figure \ref{fig:editing_comparison} demonstrates the difference between greedy search for interpolation parameters as in \citet{Edo} and per layer optimization (ours).  We are not comparing directly to \citet{Edo} since we also modified the architecture of StyleGAN to include segmentation and condition on pose, however the comparison between greedy search and our optimization is valuable.  Our method is able to preserve the shape, color, texture, and details of the region of interest (sleeve length) without affecting the other semantic regions. For example, TryOnGAN can transfer light colored pants to a person originally wearing dark jeans without lightening the rest of the image. On the other hand, TryOnGAN can change bordering regions of interest in ways that are consistent with the region of interest being transferred. For example, when transferring a short-sleeved shirt to a person with a longer-sleeved shirt, TryOnGAN synthesizes skin to show more of the arms in the final try-on image. 

\subsection{Modified StyleGAN2 Architecture Justification}
\begin{figure}
    \centering
    \includegraphics[width=0.4\textwidth]{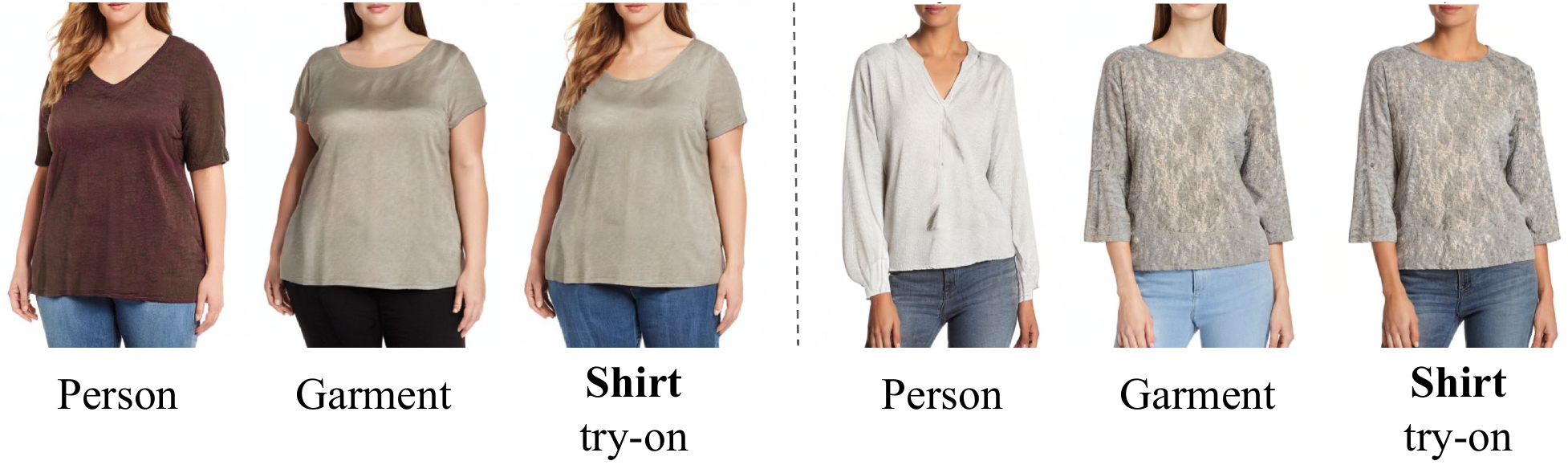}
    \caption{Try-on results for generated images using interpolation optimization with original StyleGAN2 (no pose conditioning or segmentation branch).}
    \label{fig:method_ablation}
\end{figure}

Figure \ref{fig:method_ablation} shows results on generated images using our interpolation optimization method with the original StyleGAN2 architecture (no segmentation or pose-conditioning). When $I^p$ and $I^g$ have the same pose/body type, we are able to approximate $S^t$ using the identity segments of $S^p$ and the garment segments of $S^g$. To generalize to try-on across different poses, we needed a way to segment $I^t$ during each iteration of the optimization. We made a design decision to have our network segment the generated image rather than build a third-party segmentation algorithm into our optimization method. Additionally, to control the pose separately from style and to give explicit control over the try-on image pose, we added pose-conditioning.     

%% file: 05_discussion.tex
\section{Discussion}

In this paper, we have presented a method for high quality try-on. We use the power of StyleGAN2 and show that it is possible to learn internal interpolation coefficients per layer to create a try-on experience. Our method outperforms the state of the art. We have demonstrated promising results in high resolution on a challenging task of try-on.  While promising, our method still fails in cases of extreme poses and underrepresented garments. Similarly, when projection of real images is unsatisfactory it directly affects the interpolation results, since interpolation assumes perfect projection. It is a direction for future research to improve projection of real images onto StyleGAN latent space. Our high quality try-on results on generated images show the full capability of our optimization method once projection improves. 

The try-on application is designed to visualize fashion on any person, including different skin tones, body shapes, height, weight, and so on, in the highest quality.  However, any deployment of our methods in a real-world setting would need careful attention to responsible design decisions. Such considerations could include  labeling any user-facing image that has been recomposed, and matching the distribution of people  composed into an outfit to the underlying demographics. 